\documentclass[11pt,a4paper]{article}
\usepackage[hyperref]{acl2021}
\usepackage{times}
\usepackage{multirow}
\usepackage{graphicx}
\usepackage{latexsym}

\usepackage{microtype}
\usepackage{booktabs}
\usepackage{amsmath}
\usepackage{multirow}
\usepackage{color}
\usepackage{footmisc}
\usepackage{soul}
\usepackage{arydshln}
\usepackage{graphics}
\usepackage{graphicx}
\usepackage{algorithm}
\usepackage{xurl}
\usepackage{algpseudocode}
\usepackage{url}
\definecolor{lightgreen}{HTML}{CEF6CE}
\definecolor{lightred}{HTML}{FD8F70}

\aclfinalcopy 
\def\aclpaperid{1422} 

\title{Novel Slot Detection: A Benchmark for Discovering Unknown Slot Types in the Task-Oriented Dialogue System}

\author{Yanan Wu$^{1*}$, Zhiyuan Zeng$^{1*}$, Keqing He$^{2*}$, Hong Xu$^{1}$ \\ {\bf Yuanmeng Yan$^{1}$,} {\bf Huixing Jiang$^{2}$,} {\bf Weiran Xu$^{1}$}\thanks{\ \ The first three authors contribute equally. Weiran Xu is the corresponding author.}\\
 $^1$Pattern Recognition \& Intelligent System Laboratory \\
  $^1$Beijing University of Posts and Telecommunications, Beijing, China\\
$^{2}$Meituan Group, Beijing, China\\
  \texttt{\{yanan.wu,zengzhiyuan,xuhong,yanyuanmeng,xuweiran\}@bupt.edu.cn}\\
  \texttt{\{hekeqing,jianghuixing\}@meituan.com}
  }
\date{}

\begin{document}
\maketitle
\begin{abstract}
Existing slot filling models can only recognize pre-defined in-domain slot types from a limited slot set. In the practical application, a reliable dialogue system should know what it does not know. In this paper, we introduce a new task, Novel Slot Detection (NSD), in the task-oriented dialogue system. NSD aims to discover unknown or out-of-domain slot types to strengthen the capability of a dialogue system based on in-domain training data. Besides, we construct two public NSD datasets, propose several strong NSD baselines, and establish a benchmark for future work. Finally, we conduct exhaustive experiments and qualitative analysis to comprehend key challenges and provide new guidance for future directions\footnote{\url{https://github.com/ChestnutWYN/ACL2021-Novel-Slot-Detection}}.
\end{abstract}

\section{Introduction}  
Slot filling plays a vital role to understand user queries in personal assistants such as Amazon Alexa, Apple Siri, Google Assistant, etc. It aims at identifying a sequence of tokens and extracting semantic constituents from the user queries. Given a large scale pre-collected training corpus, existing neural-based models \cite{Mesnil2015UsingRN,liu2015recurrent,Liu_2016,goo2018slot,haihong2019novel,chen2019bert,He2020MultiLevelCT,he-etal-2020-learning-tag, yan-etal-2020-adversarial,Louvan2020RecentNM,he-etal-2020-syntactic} have been actively applied to slot filling and achieved promising results. 

\begin{figure}
    \centering
    \resizebox{.50\textwidth}{!}{
    \includegraphics{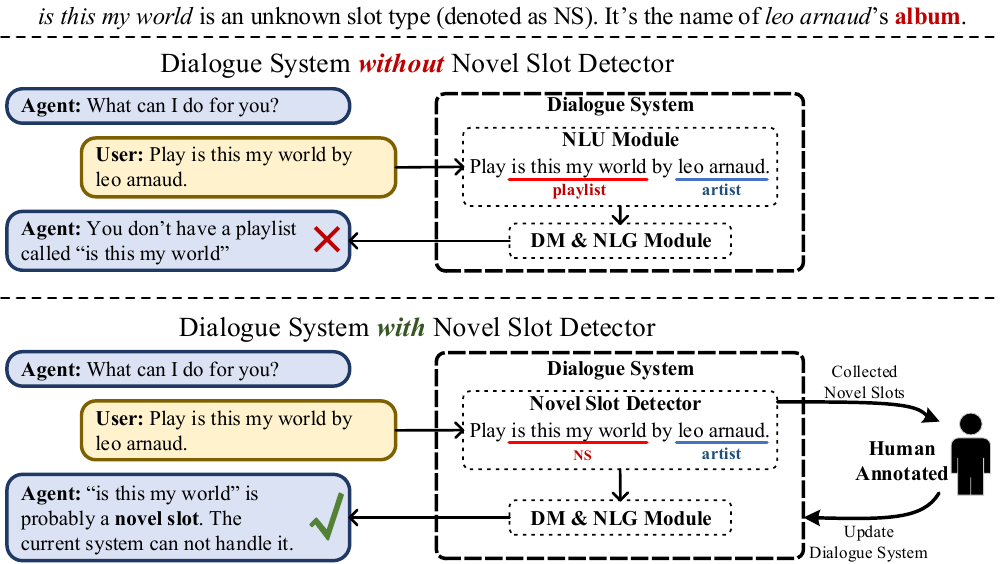}}
    \caption{An example of Novel Slot Detection in the task-oriented dialogue system. Without NSD, the dialogue system gives the wrong response since it misunderstands the unknown slot ``is this my world" as the in-domain \emph{playlist} type. In contrast, NSD recognizes ``is this my world" as \emph{NS} and the system gives a fallback response. Meanwhile, with human-in-the-loop annotation, the system can increase its functions or skills.}
    \vspace{-0.1cm}
    \label{fig:intro}
    \vspace{-0.6cm}
\end{figure}

Existing slot filling models can only recognize pre-defined entity types from a limited slot set, which is insufficient in the practical application scenario. A reliable slot filling model should not only predict the pre-defined slots but also detect potential unknown slot types to know what it doesn't know, which we call Novel Slot Detection (NSD) in this paper. NSD is particularly crucial in deployed systems—both to avoid performing the wrong action and to discover potential new entity types for future development and improvement. We display an example as Fig \ref{fig:intro} shows. 

\begin{table*}[t]
	\centering
	\resizebox{0.90\textwidth}{!}{
	\begin{tabular}{c|cccccccc}
		\hline
		Utterance & play & \textbf{is} & \textbf{this} & \textbf{my} & \textbf{world} & by & leo & arnaud \\
		\hline
		Slot Filling Labels & O & B-album & I-album & I-album & I-album & O & B-artist & I-artist \\
		Novel Slot Detection Labels & O & \textbf{NS} & \textbf{NS} & \textbf{NS} & \textbf{NS} & O & B-artist & I-artist \\
		\hline
	\end{tabular}}
	\vspace{-0.2cm}
	\caption{Comparison between slot filling and novel slot detection. In the novel slot detection labels, we consider ``album" as an unknown slot type that is out of the scope of the pre-defined slot set. Meanwhile, ``artist" belonging to in-domain slot types still needs to be recognized as the original slot filling task.}
	\label{table:atis_example}
	\vspace{-0.5cm}
\end{table*}

In this paper, we define \emph{Novel Slot} (NS) as new slot types that are not included in the pre-defined slot set. NSD aims to discover potential new or out-of-domain entity types to strengthen the capability of a dialogue system based on in-domain pre-collected training data. There are two aspects in the previous work related to NSD, out-of-vocabulary (OOV) recognition \cite{liang-etal-2017-combining,zhao-feng-2018-improving,hu-etal-2019-shot,He2020LearningLO,he-etal-2020-learning-tag,yan-etal-2020-adversarial,He2020ContrastiveZL} and out-of-domain (OOD) intent detection \cite{Lin2019DeepUI,Larson2019AnED,Xu2020ADG,9413908,Zeng2021AdversarialSL}. OOV means many slot types can have a large number of new slot values while the training set only obtains a tiny part of slot values. OOV aims to recognize unseen slot values in training set for pre-defined slot types, using character embedding \cite{liang-etal-2017-combining}, copy mechanism \cite{zhao-feng-2018-improving}, few/zero-shot learning \cite{hu-etal-2019-shot,He2020ContrastiveZL,Shah2019RobustZC}, transfer learning \cite{Chen2019TransferLF,He2020LearningLO,He2020MultiLevelCT} and background knowledge \cite{Yang2017LeveragingKB,he-etal-2020-learning-tag}, etc. Compared to OOV recognition, 
our proposed novel slot detection task focuses on detecting unknown slot types, not just unseen values. NSD faces the challenges of both OOV and no sufficient context semantics (see analysis in Section \ref{challenge}), greatly increasing the complexity of the task. Another line of related work is OOD intent detection \cite{Hendrycks2017ABF,Lee2018ASU,Lin2019DeepUI,Ren2019LikelihoodRF,Zheng2020OutofDomainDF,Xu2020ADG} which aims to know when a query falls outside the range of predefined supported intents. The main difference is that NSD detects unknown slot types in the token level while OOD intent detection identifies out-of-domain intent queries. NSD requires a deep understanding of the query context and is prone to label bias of \emph{O} (see analysis in Section \ref{data_processing}), making it challenging to identify unknown slot types in the task-oriented dialog system.

In this paper, we first introduce a new and important task, Novel Slot Detection (NSD), in the task-oriented dialogue system (Section \ref{NSD}). NSD plays a vital role in avoiding performing the wrong action and discovering potential new entity types for the future development of dialogue systems. Then, we construct two public NSD datasets, Snips-NSD and ATIS-NSD, based on the original slot filling datasets, Snips \cite{Coucke2018SnipsVP} and ATIS \cite{Hemphill1990TheAS} (Section \ref{datasets}). From the perspective of practical application, we consider three kinds of dataset construction strategies, Replace, Mask and Remove. Replace denotes we label the novel slot values with all \emph{O} in the training set. Mask is to label with all \emph{O} and mask the novel slot values. Remove is the most strict strategy where all the queries containing novel slots are removed. We dive into the details of the three different construction strategies in Section \ref{strategy} and perform a qualitative analysis in Section \ref{data_processing}. Besides, we propose two kinds of evaluation metrics, span-level F1 and token-level F1 in Section \ref{metrics}, following the slot filling task. Span F1 considers the exact matching of a novel slot span while Token F1 focuses on prediction accuracy on each word of a novel slot span. We discuss performance comparison between the two metrics and propose a new metric, restriction-oriented span evaluation (ROSE), to combine the advantages of both in Section \ref{newmetric}. Then, we establish a fair benchmark and propose extensive strong baselines for NSD in Section \ref{methods}. Finally, we perform exhaustive experiments and qualitative analysis to shed light on the challenges that current approaches faced with NSD in Section \ref{analysis} and \ref{discussion}. 

Our contributions are three-fold: (1) We introduce a Novel Slot Detection (NSD) task in the task-oriented dialogue system. NSD helps avoid performing the wrong action and discovering potential new entity types for increasing functions of dialogue systems. (2) We construct two public NSD datasets and establish a benchmark for future work. (3) We conduct exhaustive experiments and qualitative analysis to comprehend key challenges and provide new guidance for future NSD work.

\section{Problem Formulation}  
\subsection{Slot Filling}
Given a sentence $X =\{x_{1}, ..., x_{n}\}$ with $n$ tokens, the slot filling task is to predict a corresponding tag sequence $Y = \{y_{1}, ..., y_{n}\}$ in BIO format, where each $y_{i}$ can take three types of values: B-slot\_type, I-slot\_type and  \emph{O}, where ``B" and ``I" stand for the beginning and intermediate word of a slot and ``\emph{O}" means the word does not belong to any slot. Here, slot filling assumes $y_{i} \in \mathbf{y}$, where $\mathbf{y}$ denotes a pre-defined slot set of size $\mathcal{M}$. Current approaches typically model slot filling as a sequence labeling problem using RNN \cite{liu2015recurrent,Liu_2016,goo2018slot} or pre-trained language models \cite{chen2019bert}. 

\subsection{Novel Slot Detection}

\label{datasets}
\begin{table*}[t]
	\centering
	\resizebox{0.85\textwidth}{!}{
	\begin{tabular}{l|l|cccccccc}
		\hline
		\multicolumn{2}{l|}{Original Utterance} & play & \textbf{is} & \textbf{this} & \textbf{my} & \textbf{world} & by & leo & arnaud \\
		\multicolumn{2}{l|}{Original Slot Filling Labels} & O & \textbf{B-album} & \textbf{I-album} & \textbf{I-album} & \textbf{I-album} & O & B-artist & I-artist \\ \hline
		\multirow{6}{*}{Strategy} & \multirow{2}{*}{\textbf{Replace}} & play & \textbf{is} & \textbf{this} & \textbf{my} & \textbf{world} & by & leo & arnaud \\ 
		& & O & \textbf{O} & \textbf{O} & \textbf{O} & \textbf{O} & O & B-artist & I-artist \\ \cline{2-10}
		  & \multirow{2}{*}{\textbf{Mask}} & play & \textbf{MASK} & \textbf{MASK} & \textbf{MASK} & \textbf{MASK} & by & leo & arnaud \\
		  & & O & \textbf{O} & \textbf{O} & \textbf{O} & \textbf{O} & O & B-artist & I-artist \\ \cline{2-10}
		  & \multirow{2}{*}{\textbf{Remove}} & - & - & - & - & - & - & - & - \\ 
		  & &  - & - & - & - & - & - & - & - \\ 
		\hline
	\end{tabular}}
	\vspace{-0.2cm}
	\caption{Comparison between three processing strategies in the training set. We consider “album” as an unknown slot type and ``-" denotes the sentence is removed from the training data.}
	\label{table:data_strategy}
	\vspace{-0.5cm}
\end{table*}

\label{NSD}
We refer to the above training data $D$ as in-domain (IND) data. Novel slot detection aims to identify unknown or out-of-domain (OOD) slot types via IND data while correctly labeling in-domain data. We denote unknown slot type as \emph{NS} and in-domain slot types as IND in the following sections. Note that we don't distinguish between B-NS and I-NS and unify them as \emph{NS} because we empirically find existing models hardly discriminate B and I for an unknown slot type. We provide a detailed analysis in Section \ref{newmetric}. We show an example of NSD in Table \ref{table:atis_example}. The challenges of recognizing NSD come from two aspects, \emph{O} tags and in-domain slots. On the one hand, models need to learn entity information for distinguishing \emph{NS} from \emph{O} tags. On the other hand, they require discriminating \emph{NS} from other slot types in the pre-defined slot set. We provide a detailed error analysis in Section \ref{error_analysis}. 

\vspace{-0.05cm}
\section{Dataset}  
\vspace{-0.05cm}

Since there are not existing NSD datasets, we construct two new datasets based on the two widely used slot filling datasets, Snips \cite{Coucke2018SnipsVP} and ATIS \cite{Hemphill1990TheAS}. We first briefly introduce Snips and ATIS, then elaborate on data construction and processing in detail, and display the statistic of our NSD datasets, Snips-NSD and ATIS-NSD. Finally, we define two evaluation metrics for the NSD task, Span F1 and Token F1.

\subsection{Original Slot Filling Datasets}
Snips\footnote{https://github.com/sonos/nlu-benchmark/tree/master/2017-06-custom-intent-engines} is a custom intent engine dataset. It originally has 13,084 train utterances, 700 and 700 test utterances. ATIS\footnote{https://github.com/yvchen/JointSLU/tree/master/data} contains audio recordings of people making flight reservations. It originally has 4,478 train utterances, 500 dev and 893 test utterances. The full statistic is shown in Table \ref{tab:original_dataset}. Note that the vocabulary only contains words in the training set, and test set words that do not exist in the vocabulary are referred to OOV words. The percentage of OOV words represents the portion of OOV words in the test set.

\subsection{Data Construction and Processing}
\label{strategy}

For Snips and ATIS datasets, we keep some slot classes in training as unknown and integrate them back during testing, following \cite{Fei2016BreakingTC,Shu2017DOCDO,Lin2019DeepUI}. We randomly select part of slot types in Snips and ATIS as unknown slots(5\%, 15\%, and 30\% in this paper). Note that the original train/val/test split is fixed. Considering class imbalance, we perform weighted sampling where the chosen probability is relevant to the number of class examples similar to \cite{Lin2019DeepUI}. To avoid randomness of experiment results, we report the average result over 10 runs. 

\begin{table}[t]
\centering
\small
\begin{tabular}{|l|c|c|}
\hline
                     & Snips  & ATIS  \\ \hline \hline
Vocabulary Size      & 11,241   & 722 \\ \hline
Percentage of OOV words & 5.95\% & 0.77\% \\ \hline
Number of Slots      & 39   & 79     \\ \hline
Training Set Size    & 13,084 & 4,478 \\ \hline
Development Set Size & 700   & 500    \\ \hline
Testing Set Size     & 700   & 893   \\ \hline
\end{tabular}
\vspace{-0.2cm}
\caption{Statistics of ATIS and Snips datasets.}
\label{tab:original_dataset}
\vspace{-0.5cm}
\end{table}

After we choose the unknown slot types, a critical problem is how to handle sentences including these unknown slot types in training set. For OOD intent detection, we just need to remove these sentences in training and validation set. However, for Novel Slot Detection, a sentence perhaps contains both in-domain slots and unknown slots, which is nontrivial for tackling unknown slots at the token level. We need to balance the performance of recognizing unknown slots and in-domain slots. Therefore, we propose three different processing strategies as follows:
(1) \textbf{Replace}: We label the unknown slot values with all \emph{O} in the training set while the original values remain unchanged.
(2) \textbf{Mask}: We label the unknown slot values with all \emph{O} and mask these slot values with a special token \textit{MASK}. 
(3) \textbf{Remove}: All the sentences containing unknown slots are directly removed.

We display examples of the above three strategies in Table   
 \ref{table:data_strategy}. For the val and test set, we just label the unknown slot values with all \emph{NS} while keeping the in-domain labeling fixed. Note that \emph{NS} tags only exist in the val and test set, not in the training set. Besides, we keep original in-domain slots fixed to evaluate the performance of both \emph{NS} and in-domain slots. We aim to simulate the practical scenario where we can hardly know what unknown slots are. These three strategies all have its practical significance. Compared with others, Remove is the most suitable strategies for real-world scenarios. In practical scenario, dialog systems first train in the data set labeled by human annotators, and then applied to the actual application. In the process of interaction with the real users, novel slot types appear gradually. Therefore, we consider that the training set doesn't contain potential novel slots sentences. In other words, Remove is the most suitable strategy for NSD in real applications. What's more, Section \ref{data_processing} demonstrates Remove performs best while the others suffer from severe model bias by \emph{O} tags. Therefore, we adopt Remove as the main strategy in this paper.


\subsection{Statistic of New NSD Datasets}
Table \ref{Snips-NSD} shows the detailed statistics of Snips-NSD-15\% constructed by Remove strategy, where we choose 15\% classes in the training data as unknown slots. \footnote{Since different proportions of unknown slots have different statistics, here we only display the results of Snips-NSD-15\% for brevity.} Combining Table \ref{tab:original_dataset} and Table \ref{Snips-NSD}, we can find Remove strategy removes 28.70\% of queries in the original Snips training set, hence increases the percentage of OOV word from 5.95\% to 8.51\%. And unknown slot values account for 12.29\% of total slot values in the test set.

\begin{table}[t]
\centering
\resizebox{0.9\linewidth}{!}{
\begin{tabular}{|l|c|c|c|}
\hline
Snips-NSD-15\%                                                                       & Train      & Val  & Test   \\ \hline
number of in-domain slots                                                            &  33  &  33  & 33 \\  
number of unknown slots                                                              &  6 & 6 & 6     \\ 
percentage of OOV words                                                              & - & - & 8.51\% \\ \hline
number of queries                                                                    & 9,329      & 700  & 700    \\ \hline
\begin{tabular}[c]{@{}l@{}}number of queries \\ including unknown slots\end{tabular} & 0          &   192  & 202    \\ \hline
number of slot values                                                                & 23,176     &   1,794    & 1,790  \\ \hline
number of unknown slot values                                                        & 0          &   210   & 220   \\
\hline
\end{tabular}}
\caption{The detailed statistics of Snips-NSD-15\%.}
\label{Snips-NSD}
\end{table}

\subsection{Metrics}
\label{metrics}
The traditional slot filling task uses Span F1 \footnote{\url{https://www.clips.uantwerpen.be/conll2000/chunking/conlleval.txt}} for evaluation. Span F1 considers the exact span matching of an unknown slot span. However, we find in Section \ref{newmetric} that this metric is too strict to NSD models. In the practical application, we only need to coarsely mine parts of words of unknown slots, then send these queries containing potential unknown slot tokens to human annotators, which has effectively reduced extensive labor and improved efficiency. Therefore, we define a more reasonable metric, Token F1 which focuses on the word-level matching of a novel slot span. We also propose a new metric, Restriction-Oriented Span Evaluation (ROSE), for a fair comparison in Section \ref{newmetric}.

\begin{figure}
    \centering
    \resizebox{.48\textwidth}{!}{
    \includegraphics{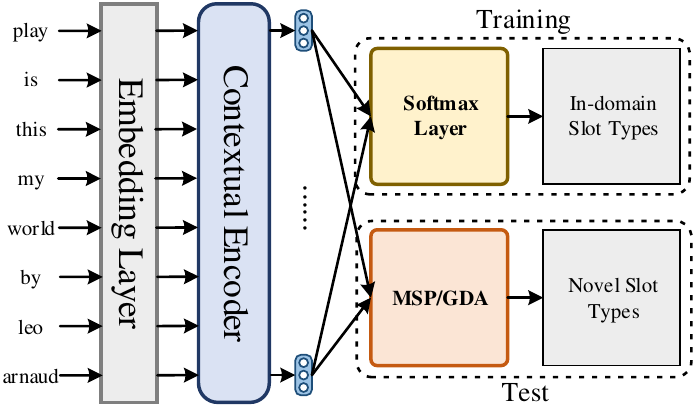}}
    \caption{The overall architecture of our approach.}
    \label{fig:model}
\end{figure}

\section{Methodology}  
\label{methods}
In this section, we introduce the NSD models proposed in this paper and illustrate the differences between the various parallel approaches during the training and test stage.

\subsection{Overall Framework}

The overall structure of model is shown in Fig \ref{fig:model}. In the training stage, we either train a multiple-class classifier or binary classifier using different training objectives. We use public BERT-large \cite{Devlin2019BERTPO} embedding layer and BiLSTM-CRF \cite{Huang2015BidirectionalLM} for token level feature extraction. Then, in the test stage, we use the typical neural multiple classifier to predict the in-domain slot labels. Meanwhile, we use the detection algorithm, MSP or GDA to figure out novel slot tokens. Finally, we override the slot token labels which are detected as NS. In terms of training objectives, detection algorithms, and distance strategies, we compare different variants as follows.

\begin{table*}[t]
\centering
\resizebox{1.0\textwidth}{!}{%
\begin{tabular}{c|c|c|c|cc||c|cc||c|cc}
\hline
\multicolumn{3}{c|}{\multirow{2}{*}{Models}} & \multicolumn{3}{c|}{5\%} & \multicolumn{3}{c|}{15\%} & \multicolumn{3}{c}{30\%} \\ \cline{4-12} 
\multicolumn{3}{l|}{} & \multicolumn{1}{c|}{IND} & \multicolumn{2}{c|}{NSD} & \multicolumn{1}{c|}{IND} & \multicolumn{2}{c|}{NSD} & \multicolumn{1}{c|}{IND} & \multicolumn{2}{c}{NSD} \\ \hline
detection method & objective & \multicolumn{1}{l|}{distance strategy} & \multicolumn{1}{l|}{Span F1} & \multicolumn{1}{l}{Span F1} & \multicolumn{1}{l|}{Token F1} & \multicolumn{1}{l|}{Span F1} & \multicolumn{1}{l}{Span F1} & \multicolumn{1}{l|}{Token F1} & \multicolumn{1}{l|}{Span F1} & \multicolumn{1}{l}{Span F1} & \multicolumn{1}{l}{Token F1} \\ \hline
\multirow{3}{*}{MSP} & binary & - & 87.21 & 12.34 & 25.16 & 71.44 & 12.31 & 39.50 & 58.88 & 8.73 & 40.38 \\ 
 & multiple & - & 88.05 & 14.04 & 30.50 & 79.71 & 20.97 & 40.02 & 78.52 & 25.26 & 46.91 \\ 
 & binary+multiple & - & 89.59 & 23.58 & 37.55 & 83.72 & 24.70 & 45.32 & 79.08 & 30.66 & 52.10 \\ \hline
\multirow{4}{*}{GDA} & binary & difference & 87.95 & 23.83 & 35.83 & 83.65 & 22.06 & 43.99 & 78.72 & 32.50 & 44.13 \\ 
 & binary & minimum & 61.29 & 10.36 & 17.08 & 49.11 & 16.91 & 31.10 & 48.07 & 15.56 & 33.78 \\ 
 & multiple & difference & \textbf{93.14} & 29.73 & 45.99 & 90.07 & 31.96 & 53.02 & 85.56 & 36.16 & 54.55 \\ 
 & multiple & minimum & 93.10 & \textbf{31.67}* & \textbf{46.97}* & \textbf{90.18} & \textbf{32.19} & \textbf{53.75}* & \textbf{86.26}* & \textbf{38.64}* & \textbf{55.24}* \\ \hline
\end{tabular}%
}
\vspace{-0.2cm}
\caption{IND and NSD results with different proportions (5\%, 15\% and 30\%) of classes are treated as unknown slots on Snips-NSD. * indicates the significant improvement over all baselines (p $<$ 0.05).}
\label{tab_main_snips}
\end{table*}

\begin{table*}[t]
\centering
\resizebox{1.0\textwidth}{!}{%
\begin{tabular}{c|c|c|c|cc||c|cc||c|cc}
\hline
\multicolumn{3}{c|}{\multirow{2}{*}{Models}} & \multicolumn{3}{c|}{5\%} & \multicolumn{3}{c|}{15\%} & \multicolumn{3}{c}{30\%} \\ \cline{4-12} 
\multicolumn{3}{l|}{} & \multicolumn{1}{c|}{IND} & \multicolumn{2}{c|}{NSD} & \multicolumn{1}{c|}{IND} & \multicolumn{2}{c|}{NSD} & \multicolumn{1}{c|}{IND} & \multicolumn{2}{c}{NSD} \\ \hline
detection method & objective & \multicolumn{1}{l|}{distance strategy} & \multicolumn{1}{l|}{Span F1} & \multicolumn{1}{l}{Span F1} & \multicolumn{1}{l|}{Token F1} & \multicolumn{1}{l|}{Span F1} & \multicolumn{1}{l}{Span F1} & \multicolumn{1}{l|}{Token F1} & \multicolumn{1}{l|}{Span F1} & \multicolumn{1}{l}{Span F1} & \multicolumn{1}{l}{Token F1} \\ \hline
\multirow{3}{*}{MSP} & binary & - & 92.04 & 19.73 & 29.63 & 91.74 & 23.40 & 33.89 & 80.49 & 21.88 & 39.17 \\
 & multiple & - & 94.33 & 27.15 & 31.16 & 92.54 & 39.88 & 42.29 & 87.63 & 40.42 & 47.64 \\ 
& binary+multiple & - & 94.41 & 32.49 & 43.48 & 93.29 & 41.23 & 43.13 & 90.14 & 41.76 & 51.87 \\ \hline
\multirow{4}{*}{GDA} & binary & difference & 93.69 & 27.02 & 34.21 & 92.13 & 30.51 & 36.30 & 88.73 & 30.91 & 45.64 \\
& binary & minimum & 93.57 & 15.90 & 20.96 & 90.98 & 24.53 & 27.26 & 88.21 & 26.40 & 39.83 \\
& multiple & difference & 95.20 & \textbf{47.78}* & \textbf{51.54}* & \textbf{93.92} & \textbf{50.92}* & \textbf{52.24}* & \textbf{92.02} & \textbf{51.26}* & \textbf{56.59}* \\
& multiple & minimum & \textbf{95.31}* & 41.74 & 45.91 & 93.88 & 43.78 & 46.18 & 91.67 & 45.44 & 52.37 \\ \hline
\end{tabular}%
}
\vspace{-0.2cm}
\caption{IND and NSD results with different proportions (5\%, 15\% and 30\%) of classes are treated as unknown slots on ATIS-NSD. * indicates the significant improvement over all baselines (p $<$ 0.05).}
\label{tab_main_atis}
\end{table*}

\noindent\textbf{Training objective.} For in-domain slots, we propose two training objectives. Multiple classifier refers to the traditional slot filling objective setting, which performs token-level multiple classifications on the BIO tags \cite{ratinov2009design} combined with different slots. Binary classifier unifies all non-O tags into one class, and the model makes a token-level binary classification of O or non-O on the sequence. Note that in the test stage, for in-domain prediction, we both use the multiple classifier. While, for novel slot detection, we use the multiple classifier, or the binary classifier, or both of them. In Table \ref{tab_main_snips} and Table \ref{tab_main_atis}, binary+multiple means the token will be labeled as NS only if both classifiers predict it as NS. 

\noindent\textbf{Detection algorithm.}
MSP and GDA are detection algorithms in the test stage. MSP (Maximum Softmax Probability) \cite{Hendrycks2017ABF} applies a threshold on the maximum softmax probability, if the maximum falls below the threshold, the token will be predicted to be a novel slot token. GDA (Gaussian Discriminant Analysis) \cite{Xu2020ADG} is a generative distance-based classifier for out-of-domain detection with Euclidean space. We treat tokens not belonging to any in-domain slots (including O) as novel slot tokens for both methods. For example, with a binary classifier, if the softmax probabilities belonging to O or non-O are both lower than an MSP threshold, then the token is labeled as NS.

\noindent\textbf{Distance strategy.}
The GDA detection is based on the distances between a target and each slot representation cluster. In original GDA, when the minimum distance is greater than a certain threshold, it is predicted to be novel slots. We propose a novel strategy named Difference, which uses the maximum distance minus the minimum distance, when the difference value of a target is less than a threshold, it is predicted as novel slots. Both of their thresholds are obtained by optimizing the NSD metrics on the validation set.

\section{Experiment and Analysis} 

\subsection{Implementation Details}
We use the public pre-trained Bert-large-uncased model to embed tokens which has 24 layers, 1024 hidden states, 16 heads and 336M parameters. The hidden size for the BiLSTM layer is set to 128. Adam is used for optimization with an initial learning rate of 2e-5. The dropout value is fixed as 0.5, and the batch size is 64. We train the model only on in-domain labeled data. The training stage has an early stopping setting with patience equal to 10. We use the best F1 scores on the validation set to calculate the MSP and GDA thresholds adaptively. Each result of the experiments is tested for 10 times under the same setting and reports the average value. The training stage of our model lasts about 28 minutes on single Tesla T4 GPU(16 GB of memory). 

\vspace{-0.05cm}
\subsection{Main Results}  
\label{mainresults}

Table \ref{tab_main_snips} and \ref{tab_main_atis} show the experiment results with seven different models on two benchmark slot filling datasets Snips-NSD and ATIS-NSD constructed by Remove strategy. We both report NSD and IND results using Span F1 and Token F1. We compare these models from three perspectives, detection method, objective and distance strategy in the following. The analysis of effect of the proportion of unknown slot types is described in \ref{split}.


\noindent\textbf{Detection Method: MSP vs GDA.} Under the same setting of objective, GDA performs better than MSP in both IND and NSD, especially in NSD. We argue that GDA models the posterior distribution on representation spaces of the feature extractor and avoids the issue of overconfident predictions \cite{Guo2017OnCO,Liang2017PrincipledDO,Liang2018EnhancingTR}. Besides, comparing Snips-NSD and ATIS-NSD, NSD Token F1 scores on ATIS-NSD are much higher than Snips-NSD but no significant difference exists for NSD Span F1 scores. The reason is that Snips-NSD has a higher average entity length (1.83) than ATIS-NSD (1.29), making it harder to detect the exact \emph{NS} span.


\noindent\textbf{Objective: Binary vs Multiple.} Under all settings, Multiple outperforms Binary with a large margin on two datasets in both IND and NSD metrics. For MSP, combining Multiple and Binary get higher F1 scores. Specifically, the Binary classifier is used to calculate the confidence of a token belonging to non-\emph{O} type, which can judge whether the token belongs to entities and distinguish \emph{NS} from type \emph{O}. On the other hand, we use the Multiple classifier to calculate the confidence for tokens that are of type \emph{NS}, to distinguish \emph{NS} from all predefined non-\emph{O} slot types. For GDA, we do not combine Multiple and Binary because of poor performance. Multiple achieves the best results for all the IND and NSD F1 scores. We suppose multi-class classification can better capture semantic features than binary classification.

\begin{table}[t]
\centering
\resizebox{0.50\textwidth}{!}{%
\begin{tabular}{l|l|ll|l|ll|l|ll}
\hline
\multirow{3}{*}{Strategy} & \multicolumn{3}{c|}{5\%} & \multicolumn{3}{c|}{15\%} & \multicolumn{3}{c}{30\%} \\ \cline{2-10} 
 & IND & \multicolumn{2}{c|}{NSD} & IND & \multicolumn{2}{c|}{NSD} & IND & \multicolumn{2}{c}{NSD} \\ \cline{2-10} 
 & Span& Span& Token& Span& Span& Token & Span & Span & Token \\ \hline
Replace & 94.52 & 1.93 & 5.27 & 94.33 & 0.66 & 2.29 & 94.02 & 0.27 & 0.82 \\ \hline
Mask & 90.08 & 23.10 & 37.91 & 86.52 & 25.07 & 45.92 & 83.37 & 32.14 & 50.68 \\ \hline
Remove & 93.10 & \textbf{31.67} & \textbf{46.97} & 90.18 & \textbf{32.19} & \textbf{53.75} & 86.26 & \textbf{38.64} & \textbf{55.24} \\ \hline
\end{tabular}%
}
\vspace{-0.1cm}
\caption{Comparison between different data processing strategies on Snips-NSD using GDA+Multiple+Minimum.}
\label{tab_ana1}
\end{table}

\begin{figure}[t]
    \flushleft
    \resizebox{.52\textwidth}{!}{
    \includegraphics{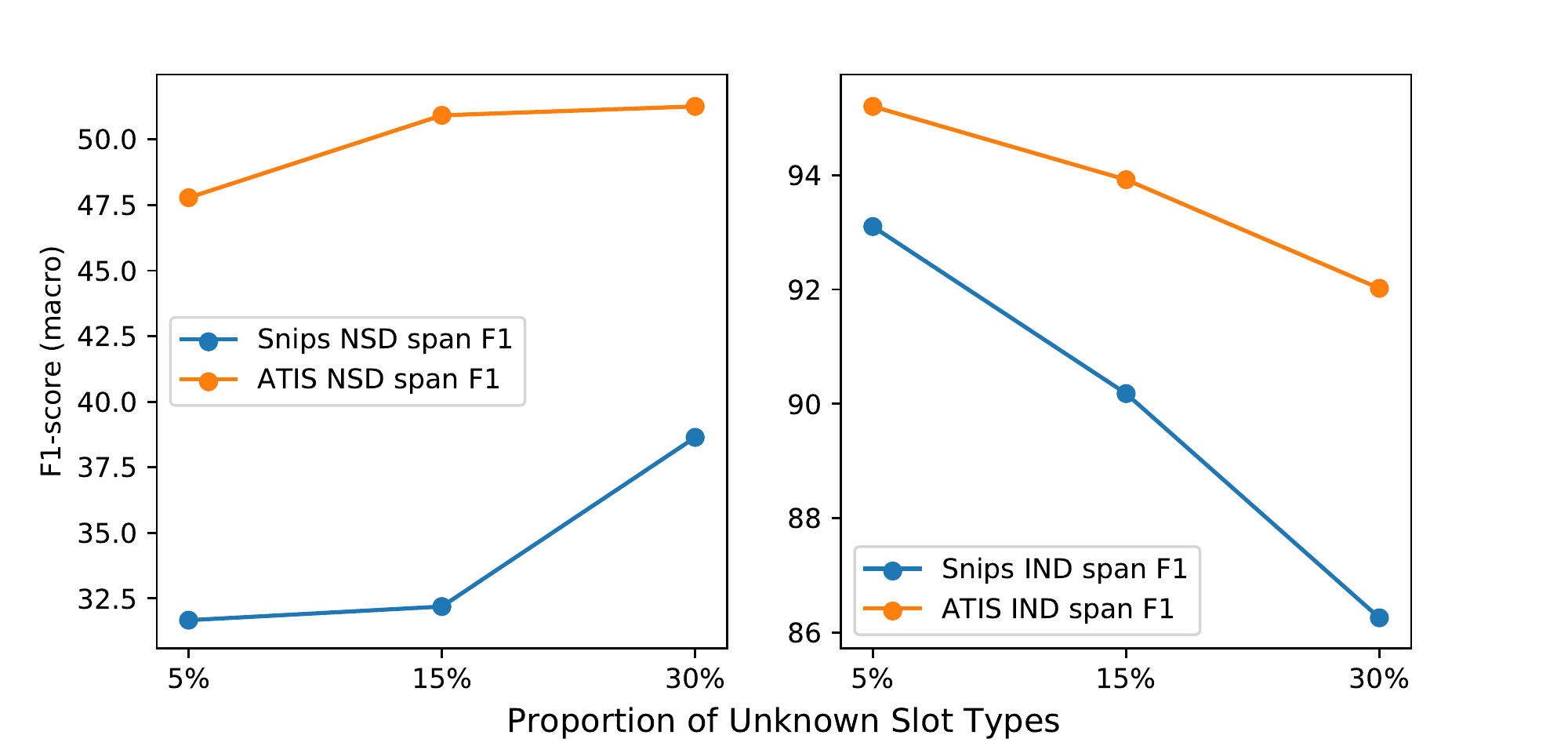}}
    \vspace{-0.7cm}
    \caption{Effect of the proportion of unknown slot types.}
    \label{fig:propotion}
     \vspace{-0.4cm}
\end{figure}

\noindent\textbf{Distance Strategy: Minimum vs Difference.} We find under the same setting of Binary, Difference strategy outperforms Minimum on both datasets for NSD metrics. But under the same setting of Multiple, there is no consistent superiority between the two distance strategies. For example, Difference outperforms Minimum for NSD metrics on ATIS-NSD, opposite to the results on Snips-NSD. We argue different distance strategies are closely related to objective settings and dataset complexity. We will leave the theoretical analysis to the future.

\subsection{Qualitative Analysis}  
\label{analysis}

\subsubsection{Effect of Different Data Processing Strategies}
\label{data_processing}

Table \ref{tab_ana1} displays IND and NSD metrics of three different dataset processing strategies on Snips-NSD using the same model GDA+Multiple+Minimum. In this section, we will dive into the analysis of the effects of different data processing strategies. Results show the Replace strategy gets poor performance in NSD, which proves labeling unknown slots as \emph{O} tags will severely mislead the model. The Mask and Remove strategies are more reasonable since they remove unknown slots from the training data. Their main difference is that Mask only deletes token-level information, while Remove even eliminates the contextual information. For NSD in all datasets, Remove gains significantly better performance on both Token F1 and Span F1 than Mask by 9.06\%(5\%), 7.83\%(15\%) and 4.56\%(30\%) on Token F1, and 8.57\%(5\%), 7.12\%(15\%) and 6.5\%(30\%) on Span F1. We argue the remaining context is still misleading even if the novel slot tokens are not directly trained in the Mask strategy. Besides, Mask does not conform to the real NSD scenario. Generally, Remove is the most suitable strategy for NSD in real applications and can achieve the best performance.

\vspace{-0.05cm}
\subsubsection{Effect of the Proportion of Unknown Slot Types}
\vspace{-0.05cm}
\label{split}

Fig \ref{fig:propotion} displays the effect of the proportion of unknown slot types using the Remove strategy in GDA+Multiple+Minimum. Results show that with the increase of the proportion of unknown slot types, the NSD F1 scores get improvements while IND F1 scores decrease. We suppose fewer in-domain slot types help the model distinguish unknown slots from IND slots, thus NSD F1 scores get improvements. However, for in-domain slot detection, since Remove deletes all the sentences containing unknown slots in the training data, our models suffer from the lack of sufficient context to recognize IND slots so IND F1 scores decrease.


\begin{figure}[t]
    \centering
    \vspace{-0.35cm}
    \resizebox{.50\textwidth}{!}{
    \includegraphics{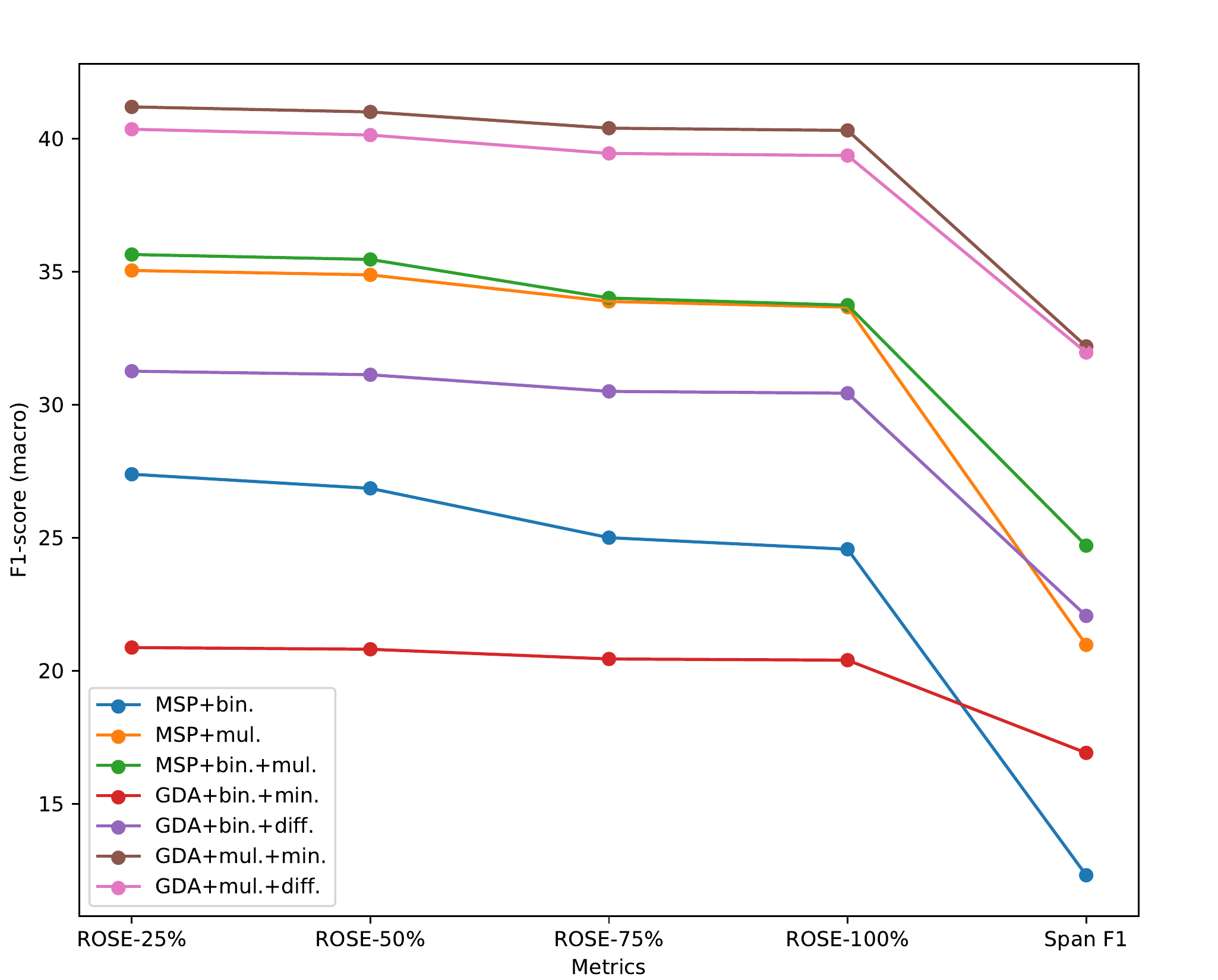}}
    \vspace{-0.3cm}
    \caption{Effect of varying degrees of restrictions}
    \label{fig_metric}
\end{figure}

\begin{table}[t]
\centering
\resizebox{0.45\textwidth}{!}{%
\begin{tabular}{l|c|c}
\hline
 & GDA+mul.+min. & MSP+bin.+mul. \\ \hline
ROSE-mean & 40.73 & 34.71 \\ \hline
ROSE-100\% & 40.39 & 33.74 \\ \hline
ROSE-50\% & 41.00 & 35.46 \\ \hline
\end{tabular}%
}
\vspace{-0.2cm}
\caption{ROSE metrics on Snips-NSD using GDA+Multiple+Minimum and MSP+Binary+Multiple}
\vspace{-0.2cm}
\label{tab_rosemean}
\end{table}

\subsubsection{New Metric: ROSE}
\label{newmetric}

The previous results have shown Span F1 is much lower than the token F1. The reason is that Span F1 is a strict metric, where the model needs to correctly predict all \emph{NS} tokens and the correct boundary. This is difficult for NSD models due to the lack of supervised information. In fact, NSD models only need to mark some tokens in the span of novel slots and send the total sequence containing the \emph{NS} tokens back to the humans. A small number of token omissions or misjudgments are acceptable. Therefore, to meet a reasonable NSD scenario, we propose a new metric, restriction-oriented span evaluation (ROSE), to evaluate the span prediction performance under different restrictions. First, we do not punish the situation where tokens prediction exceeds the span. Then, we consider a span is correct when the number of correctly predicted tokens is greater than a settable proportion $p$ of the span length. We take the average of the ROSE score and the original span F1 to avoid the model obtaining an outstanding result through over-long prediction. The results using Snips with 15\% of novel slots are shown in Figure \ref{fig_metric}. As the degree of restriction increases, the metrics tend to decline. It indicates that the model can mostly identify more than half of the tokens in spans. To make a comprehensive evaluation, we defined the ROSE-mean, namely the mean of ROSE-25\%, ROSE-50\%, ROSE-75\%, and ROSE-100\%. We present results on part of proposed models in Table \ref{tab_rosemean}.

\subsubsection{Analysis of Single Unknown Slot}
\label{ana_single}

To analyze the relationship between NSD performance and a single specific slot, we calculate the token and span metrics treating each single slot type as an unknown slot and show the results of the top five and bottom five for Token F1 scores in Table \ref{tab_ana_2}. We find that the slots with better performance often account for a larger percentage of the data set, such as Object\_name or Entity\_name. They also tend to have a larger value space, such as TimeRange, Music\_item, or Artist. These characteristics allow the semantic representation of these slots to be distributed over a large area rather than clustered tightly together. We consider that this distribution is more reasonable because in a real application scenario, novel slots are diverse and its distribution tends to be diffuse. Performance on these types also proves that the NSD models we propose can be better generalized to a reasonable data setting.

\begin{table}[]
\centering
\resizebox{0.50\textwidth}{!}{%
\begin{tabular}{c|c|c|c|c|c}
\hline
\multicolumn{2}{c|}{Type} & Proportion(\%) & Span Length & Token F1 & Span F1 \\ \hline
\multirow{5}{*}{top 5} & Object\_name & 21.42 & 3.71 & 55.64 & 20.82 \\ \cline{2-6} 
 & TimeRange & 15.29 & 2.35 & 53.65 & 30.15 \\ \cline{2-6} 
 & Entity\_name & 23.14 & 3.09 & 48.56 & 22.83 \\ \cline{2-6} 
 & Music\_item & 14.86 & 1.05 & 46.23 & 34.59 \\ \cline{2-6} 
 & Artist & 15.29 & 2.05 & 45.26 & 26.36 \\ \hline
\multirow{5}{*}{bottom 5} & City & 8.57 & 1.32 & 18.72 & 15.85 \\ \cline{2-6} 
 & Country & 6.29 & 1.57 & 14.19 & 11.11 \\ \cline{2-6} 
 & State & 5.54 & 1.10 & 13.55 & 10.83 \\ \cline{2-6} 
 & Best\_rating & 6.14 & 1.00 & 11.04 & 11.04 \\ \cline{2-6} 
 & Year & 3.43 & 1.00 & 10.24 & 10.24 \\ \hline
\end{tabular}%
}
\vspace{-0.2cm}
\caption{Results of single unknown slot.}
\label{tab_ana_2}
\end{table}

\begin{table}[t]
\centering
\resizebox{0.50\textwidth}{!}{%
\begin{tabular}{l|l|c|c}
\hline
Type 1 & Type 2 & Token F1 & Span F1 \\ \hline
Object\_name & - & 55.64 & 20.82 \\ \hline
TimeRange & - & 53.65 & 30.15 \\ \hline
Party\_size\_number & - & 33.44 & 28.57 \\ \hline
City & - & 18.72 & 15.85 \\ \hline
State & - & 13.55 & 10.83 \\ \hline
Object\_name & TimeRange & 53.88 & 23.37 \\ \hline
Object\_name & Party\_size\_number & 52.81 & 22.35 \\ \hline
Object\_name & City & 57.92 & 21.42 \\ \hline
Object\_name & State & 56.32 & 19.27 \\ \hline
TimeRange & Party\_size\_number & 71.27$^*$ & 51.03$^*$ \\ \hline
City & State & 29.33$^*$ & 27.14$^*$ \\ \hline
\end{tabular}%
}
\vspace{-0.2cm}
\caption{Results of combining multiple unknown slots. * denotes that NSD performance of the combination of two unknown slots is significantly better than each single slot.}
\vspace{-0.4cm}
\label{combiningslot}
\end{table}

\subsubsection{Analysis for Relationship of Multiple Unknown Slots}
\label{ana_combining}
In order to explore the effect of inter-slot relationships on NSD, we conducted experiments in which two types are mixed as novel slots. Some of the results are shown in Table \ref{combiningslot}. In the five types shown in the table, Object\_name is an open vocabulary slot with a wide range of values and contains many OOV tokens, TimeRange and Party\_size\_number often contain numbers, City and State are usually similar in semantics and context. We found that when the other types combined with Object\_name, NSD performance is often maintained close to treat Object\_name as a novel slot alone. The reason, on the one hand, is that the proportion of other types in the dataset is relatively small, so the overall impact on the metrics is smaller. On the other hand, due to the large semantic distribution range of the open vocabulary slot, there is a latent inclusion relationship for other types, so the mixing of a single type tends to have a slight impact on the NSD performance. We also found that the appropriate combination can significantly improve the efficiency of NSD. Such as TimeRange with Party\_size\_number, or City with State. This indicates that when the novel slot is similar to the in-domain slot, the model tends to predict the novel slot as a similar slot, which leads to errors. When both are treated as novel slots, these errors can be mitigated.

\begin{table}[t]
\centering
\resizebox{0.50\textwidth}{!}{%
\begin{tabular}{l|c|c|c||c}
\hline
NSD error proportion(\%) & O & Open vocabulary slots & Other slots & Sum  \\ \hline
Prediction is NS & 17.79 & 18.84 & 9.07 & 45.70 \\ \hline
Target is NS & 18.47 & 7.54 & 28.29 & 54.30 \\ \hline\hline
 Sum & 36.26 & 26.38 & 37.36 &  100.00\\ \hline
\end{tabular}%
}
\vspace{-0.15cm}
\caption{Relative proportions of several types of errors.}
\label{errorproportion}
\vspace{-0.3cm}
\end{table}

\begin{table}[t]
\resizebox{0.50\textwidth}{!}{%
\begin{tabular}{l|l|l}
\hline
Error type & NS & Example \\ \hline
NS to O & \begin{tabular}[c]{@{}l@{}}movie\_name\\ (m\_name)\end{tabular} & \begin{tabular}[c]{@{}l@{}}text: when will paris by night aired\\ true: O O B-m\_name \textbf{I-m\_name} I-m\_name O\\ predict: O O NS \textbf{O} NS O\end{tabular} \\ \hline
\begin{tabular}[c]{@{}l@{}}NS to\\ open slot\end{tabular} & album & \begin{tabular}[c]{@{}l@{}}text: play the insoc ep\\ true: O \textbf{B-album} \textbf{I-album} I-album\\ predict: O \textbf{B-object\_name} \textbf{I-object\_name} NS\end{tabular} \\ \hline
\begin{tabular}[c]{@{}l@{}}NS to\\ other slot\end{tabular} & artist & \begin{tabular}[c]{@{}l@{}}text: play kurt cobain ballad tunes\\ true: O \textbf{B-artist} \textbf{I-artist} B-music\_item O\\ predict: O \textbf{B-genre} \textbf{I-genre} B-music\_item O\end{tabular} \\ \hline
O to NS & artist & \begin{tabular}[c]{@{}l@{}}text: the workout playlist needs more chris cross\\ true: O B-playlist O O \textbf{O} B-artist I-artist\\ predict: O B-playlist O O \textbf{NS} NS NS\end{tabular} \\ \hline
\begin{tabular}[c]{@{}l@{}}open slots\\ to NS\end{tabular} & object\_type & \begin{tabular}[c]{@{}l@{}}text: tell me the actors of the saga awards\\ true: O O O \textbf{B-object\_name} O O B-object\_type O\\ predict: O O O \textbf{NS} O O NS O\end{tabular} \\ \hline
\begin{tabular}[c]{@{}l@{}}other slots\\ to NS\end{tabular} & city & \begin{tabular}[c]{@{}l@{}}text: what is the weather of east portal ks\\ true: O O O O O B-city I-city \textbf{B-state}\\ predict: O O O O O NS NS \textbf{NS}\end{tabular} \\ \hline
\end{tabular}%
}
\vspace{-0.2cm}
\caption{Error case from NSD prediction.}
\label{errorcase}
\vspace{-0.5cm}
\end{table}

\section{Discussion}  
\label{discussion}
In this section, we empirically divide all the error samples into three categories. Each type of problem contains two aspects, corresponding to NSD precision and recall, respectively. We present the relative proportions of several types of errors in Table \ref{errorproportion}, which using Snips dataset with 5\% novel slots on GDA+multiple+minimum model. For each error type, we present an example in Table \ref{errorcase} to describe the characteristics and analyze the causes. Then, we dive into identifying the key challenges and finally proposed possible solutions for future work.

\subsection{Error Analysis}
\label{error_analysis}
\textbf{Tag \emph{O}.} Tag \emph{O} is the largest and most widely distributed type in the dataset, and it generally refers to the independent function tokens. Therefore, when identifying, it is easy to be confused with other types, and the confusion is more serious for novel slots without supervised learning. We observed that tokens with \emph{O} label detected as novel slots usually exist near spans, and the function words in the span labeled as a novel slot have a probability of being predicted as \emph{O}. We consider that this kind of problem is related to the context. Although the processing strategy of Remove can effectively reduce the misleading of \emph{O} for the novel slots, tag \emph{O} will still be affected by context information of other in-domain slots. 

\noindent \textbf{Open Vocabulary Slots.} We observe that a large number of novel slot tokens are mispredicted as open vocabulary slots, while the reverse situation is much less likely to happen. This indicates that in Snips, open vocabulary slots tend to overlap or contain most other slots semantically. Even in traditional slot filling tasks, open vocabulary slots are often confused with other slots. We demonstrate this hypothesis in the analysis. Section \ref{ana_combining} shows that NSD performs better when open vocabulary slots are treated as novel slots, and Section \ref{ana_single} shows that there is no significant performance change when open vocabulary slots are mixed with some semantically concentrated slots. The reason for this problem is that the definition of the dataset is not reasonable. Slots with a large value range can hardly help the personal assistant to give an appropriate reply, and the supervised information of these slots is usually incomplete.

\noindent \textbf{Similar Slots.} Except for the two cases mentioned above, predicting novel slots as other in-domain slots is the most common type of error, in which similar slots account for a large part of it. Due to the overlap between vocabulary or shared similar context, the model often tend to be overconfident to predict similar slot labels, we analyze the phenomenon in Table \ref{combiningslot}, when similar types is treated as a new slot at the same time, NSD efficiency will rise significantly. We employ a generative classification method GDA, compared with the traditional MSP method, to make full use of data features and alleviate the problem. 

\subsection{Challenges}
\label{challenge}
Based on the above analysis, we summarize the current challenges faced by the NSD task:

\noindent\textbf{Function tokens.} Articles, prepositions, and so on that act as connective words in a sequence. It is usually labeled with type \emph{O}, but also found in some long-span slots, such as Movie\_name. It can lead to confusion between \emph{O} and novel slot when this kind of slot is the target of NSD.

\noindent\textbf{Insufficient context.} Correct slot detection often depends on the context, and this supervised information is missing for novel slots. Models can only conduct NSD to tokens using the original embeddings or representations trained in other contexts, which can lead to bias in the semantic modeling of the novel slot.

\noindent\textbf{Dependencies between slots.} There are some semantic overlaps or inclusion relationships in the slot definition of the current benchmark slot filling datasets. As a result, the semantic features are not sufficiently discriminative, and thus some outliers tokens in in-domain slots are easily confused with the novel slots.

\noindent\textbf{Open vocabulary slots.} Open vocabulary slots is a special kind of slot, its definition is usually macroscopic and can be further divided, the value range is broad. The representation distribution for Open vocabulary slots tends to be diffuse and uneven, which can be misleading to NSD.

\subsection{Future Directions}

For tag \emph{O}, a possible solution is to use a binary model to assist identification between \emph{O} and non-\emph{O} function tokens, we provide a simple method in this paper and leave further optimizing to future work.
Then, to decouple the dependencies between slots, it is critical to learn more discriminative features for in-domain data, using contrastive learning or prototypical network is expected to help. Besides, in the traditional slot filling task, the open vocabulary slot problem has been researched for a long time, and accumulate many achievements. Adaptive combination and improvement of relevant methods with NSD tasks is also an important direction of our future research.

\section{Related Work}

\textbf{OOV Recognition} OOV aims to recognize unseen slot values in training set for pre-defined slot types, using character embedding \cite{liang-etal-2017-combining}, copy mechanism \cite{zhao-feng-2018-improving}, few/zero-shot learning \cite{hu-etal-2019-shot,Shah2019RobustZC}, transfer learning \cite{Chen2019TransferLF,He2020LearningLO} and background knowledge \cite{Yang2017LeveragingKB,he-etal-2020-learning-tag}, etc. Our proposed NSD task focuses on detecting unknown slot types, not just unseen values. 

\textbf{OOD Intent Detection} \newcite{Lee2018ASU,Lin2019DeepUI,Xu2020ADG} aim to know when a query falls outside the range of predefined supported intents. Generally, they first learn discriminative intent representations via in-domain (IND) data, then employs detecting algorithms, such as Maximum Softmax Probability (MSP) \cite{Hendrycks2017ABF},  Local Outlier Factor (LOF) \cite{Lin2019DeepUI},  Gaussian Discriminant Analysis (GDA) \cite{xu-etal-2020-deep} to compute the similarity of features between OOD samples and IND samples. Compared to our proposed NSD, the main difference is that NSD detects unknown slot types in the token level  while  OOD  intent  detection  identifies sentence-level OOD intent queries.

\section{Conclusion}

In this paper, we defined a new task, Novel Slot Detection(NSD), then provide two public datasets and establish a benchmark for it. Further, we analyze the problems of NSD through multi-angle experiments and extract the key challenges of the task. We provide some strong models for these problems and offer possible solutions for future work.

\section*{Acknowledgements}
This work was partially supported by National Key R\&D Program of China No. 2019YFF0303300 and Subject II  No. 2019YFF0303302, DOCOMO Beijing Communications Laboratories Co., Ltd, MoE-CMCC "Artifical Intelligence" Project No. MCM20190701.

\section*{Broader Impact}
Dialog systems have demonstrated remarkable performance across a wide range of applications, with the promise of a significant positive impact on human production mode and lifeway. The first step of the dialog system is to identify users' key points. In practical industrial scenario, users may make unreasonable queries which fall outside of the scope of the system-supported slot types. Previous dialogue systems will ignore this problem, which will lead to wrong operations and limit the system's development. In this paper, we firstly propose to detect not only pre-defined slot types but also potential unknown or out-of-domain slot types using MSP and GDA methods. According to exhaustive experiments and qualitative analysis, we also discuss several major challenges in Novel Slot Detection for future work. The effectiveness and robustness of the model are significantly improved by adding Novel Slot Detection, which takes a step towards the ultimate goal of enabling the safe real-world deployment of dialog systems in safety-critical domains. The experimental results have been reported on standard benchmark datasets for considerations of reproducible research.


\bibliographystyle{acl_natbib}
\bibliography{anthology,acl2021}

\begin{thebibliography}{39}
\expandafter\ifx\csname natexlab\endcsname\relax\def\natexlab#1{#1}\fi

\bibitem[{Chen and Moschitti(2019)}]{Chen2019TransferLF}
Lingzhen Chen and Alessandro Moschitti. 2019.
\newblock Transfer learning for sequence labeling using source model and target
  data.
\newblock \emph{ArXiv}, abs/1902.05309.

\bibitem[{Chen et~al.(2019)Chen, Zhuo, and Wang}]{chen2019bert}
Qian Chen, Zhu Zhuo, and Wen Wang. 2019.
\newblock Bert for joint intent classification and slot filling.
\newblock \emph{arXiv preprint arXiv:1902.10909}.

\bibitem[{Coucke et~al.(2018)Coucke, Saade, Ball, Bluche, Caulier, Leroy,
  Doumouro, Gisselbrecht, Caltagirone, Lavril, Primet, and
  Dureau}]{Coucke2018SnipsVP}
A.~Coucke, A.~Saade, Adrien Ball, Th{\'e}odore Bluche, A.~Caulier, D.~Leroy,
  Cl{\'e}ment Doumouro, Thibault Gisselbrecht, F.~Caltagirone, Thibaut Lavril,
  Ma{\"e}l Primet, and J.~Dureau. 2018.
\newblock Snips voice platform: an embedded spoken language understanding
  system for private-by-design voice interfaces.
\newblock \emph{ArXiv}, abs/1805.10190.

\bibitem[{Devlin et~al.(2019)Devlin, Chang, Lee, and
  Toutanova}]{Devlin2019BERTPO}
J.~Devlin, Ming-Wei Chang, Kenton Lee, and Kristina Toutanova. 2019.
\newblock Bert: Pre-training of deep bidirectional transformers for language
  understanding.
\newblock In \emph{NAACL-HLT}.

\bibitem[{Fei and Liu(2016)}]{Fei2016BreakingTC}
Geli Fei and B.~Liu. 2016.
\newblock Breaking the closed world assumption in text classification.
\newblock In \emph{HLT-NAACL}.

\bibitem[{Goo et~al.(2018)Goo, Gao, Hsu, Huo, Chen, Hsu, and
  Chen}]{goo2018slot}
Chih-Wen Goo, Guang Gao, Yun-Kai Hsu, Chih-Li Huo, Tsung-Chieh Chen, Keng-Wei
  Hsu, and Yun-Nung Chen. 2018.
\newblock Slot-gated modeling for joint slot filling and intent prediction.
\newblock In \emph{Proceedings of the 2018 Conference of the North American
  Chapter of the Association for Computational Linguistics: Human Language
  Technologies, Volume 2 (Short Papers)}, pages 753--757.

\bibitem[{Guo et~al.(2017)Guo, Pleiss, Sun, and Weinberger}]{Guo2017OnCO}
Chuan Guo, Geoff Pleiss, Yu~Sun, and Kilian~Q. Weinberger. 2017.
\newblock On calibration of modern neural networks.
\newblock In \emph{ICML}.

\bibitem[{Haihong et~al.(2019)Haihong, Niu, Chen, and Song}]{haihong2019novel}
E~Haihong, Peiqing Niu, Zhongfu Chen, and Meina Song. 2019.
\newblock A novel bi-directional interrelated model for joint intent detection
  and slot filling.
\newblock In \emph{Proceedings of the 57th Annual Meeting of the Association
  for Computational Linguistics}, pages 5467--5471.

\bibitem[{He et~al.(2020{\natexlab{a}})He, Lei, Yang, Jiang, and
  Wang}]{he-etal-2020-syntactic}
Keqing He, Shuyu Lei, Yushu Yang, Huixing Jiang, and Zhongyuan Wang.
  2020{\natexlab{a}}.
\newblock \href {https://www.aclweb.org/anthology/2020.coling-main.246}
  {Syntactic graph convolutional network for spoken language understanding}.
\newblock In \emph{Proceedings of the 28th International Conference on
  Computational Linguistics}, pages 2728--2738, Barcelona, Spain (Online).
  International Committee on Computational Linguistics.

\bibitem[{He et~al.(2020{\natexlab{b}})He, Xu, and Yan}]{He2020MultiLevelCT}
Keqing He, Weiran Xu, and Yuanmeng Yan. 2020{\natexlab{b}}.
\newblock Multi-level cross-lingual transfer learning with language shared and
  specific knowledge for spoken language understanding.
\newblock \emph{IEEE Access}, 8:29407--29416.

\bibitem[{He et~al.(2020{\natexlab{c}})He, Yan, hong Liu, Liu, and
  Xu}]{He2020LearningLO}
Keqing He, Yuanmeng Yan, Si~hong Liu, Z.~Liu, and Weiran Xu.
  2020{\natexlab{c}}.
\newblock Learning label-relational output structure for adaptive sequence
  labeling.
\newblock \emph{2020 International Joint Conference on Neural Networks
  (IJCNN)}, pages 1--8.

\bibitem[{He et~al.(2020{\natexlab{d}})He, Yan, and
  Xu}]{he-etal-2020-learning-tag}
Keqing He, Yuanmeng Yan, and Weiran Xu. 2020{\natexlab{d}}.
\newblock \href {https://doi.org/10.18653/v1/2020.acl-main.58} {Learning to tag
  {OOV} tokens by integrating contextual representation and background
  knowledge}.
\newblock In \emph{Proceedings of the 58th Annual Meeting of the Association
  for Computational Linguistics}, pages 619--624, Online. Association for
  Computational Linguistics.

\bibitem[{He et~al.(2020{\natexlab{e}})He, Zhang, Yan, Xu, Niu, and
  Zhou}]{He2020ContrastiveZL}
Keqing He, Jinchao Zhang, Yuanmeng Yan, Weiran Xu, Cheng Niu, and Jie Zhou.
  2020{\natexlab{e}}.
\newblock Contrastive zero-shot learning for cross-domain slot filling with
  adversarial attack.
\newblock In \emph{COLING}.

\bibitem[{Hemphill et~al.(1990)Hemphill, Godfrey, and
  Doddington}]{Hemphill1990TheAS}
C.~T. Hemphill, J.~J. Godfrey, and G.~Doddington. 1990.
\newblock The atis spoken language systems pilot corpus.
\newblock In \emph{HLT}.

\bibitem[{Hendrycks and Gimpel(2017)}]{Hendrycks2017ABF}
Dan Hendrycks and Kevin Gimpel. 2017.
\newblock A baseline for detecting misclassified and out-of-distribution
  examples in neural networks.
\newblock \emph{ArXiv}, abs/1610.02136.

\bibitem[{Hu et~al.(2019)Hu, Chen, Chang, and Sun}]{hu-etal-2019-shot}
Ziniu Hu, Ting Chen, Kai-Wei Chang, and Yizhou Sun. 2019.
\newblock \href {https://doi.org/10.18653/v1/P19-1402} {Few-shot representation
  learning for out-of-vocabulary words}.
\newblock In \emph{Proceedings of the 57th Annual Meeting of the Association
  for Computational Linguistics}, pages 4102--4112, Florence, Italy.
  Association for Computational Linguistics.

\bibitem[{Huang et~al.(2015)Huang, Xu, and Yu}]{Huang2015BidirectionalLM}
Zhiheng Huang, W.~Xu, and Kai Yu. 2015.
\newblock Bidirectional lstm-crf models for sequence tagging.
\newblock \emph{ArXiv}, abs/1508.01991.

\bibitem[{Larson et~al.(2019)Larson, Mahendran, Peper, Clarke, Lee, Hill,
  Kummerfeld, Leach, Laurenzano, Tang, and Mars}]{Larson2019AnED}
Stefan Larson, Anish Mahendran, Joseph Peper, Christopher Clarke, Andrew Lee,
  P.~Hill, Jonathan~K. Kummerfeld, Kevin Leach, M.~Laurenzano, L.~Tang, and
  J.~Mars. 2019.
\newblock An evaluation dataset for intent classification and out-of-scope
  prediction.
\newblock \emph{ArXiv}, abs/1909.02027.

\bibitem[{Lee et~al.(2018)Lee, Lee, Lee, and Shin}]{Lee2018ASU}
Kimin Lee, Kibok Lee, H.~Lee, and Jinwoo Shin. 2018.
\newblock A simple unified framework for detecting out-of-distribution samples
  and adversarial attacks.
\newblock In \emph{NeurIPS}.

\bibitem[{Liang et~al.(2017{\natexlab{a}})Liang, Xu, and
  Zhao}]{liang-etal-2017-combining}
Dongyun Liang, Weiran Xu, and Yinge Zhao. 2017{\natexlab{a}}.
\newblock \href {https://doi.org/10.18653/v1/W17-2606} {Combining word-level
  and character-level representations for relation classification of informal
  text}.
\newblock In \emph{Proceedings of the 2nd Workshop on Representation Learning
  for {NLP}}, pages 43--47, Vancouver, Canada. Association for Computational
  Linguistics.

\bibitem[{Liang et~al.(2017{\natexlab{b}})Liang, Li, and
  Srikant}]{Liang2017PrincipledDO}
Shiyu Liang, Yixuan Li, and R.~Srikant. 2017{\natexlab{b}}.
\newblock Principled detection of out-of-distribution examples in neural
  networks.
\newblock \emph{ArXiv}, abs/1706.02690.

\bibitem[{Liang et~al.(2018)Liang, Li, and Srikant}]{Liang2018EnhancingTR}
Shiyu Liang, Yixuan Li, and R.~Srikant. 2018.
\newblock Enhancing the reliability of out-of-distribution image detection in
  neural networks.
\newblock \emph{arXiv: Learning}.

\bibitem[{Lin and Xu(2019)}]{Lin2019DeepUI}
Ting-En Lin and H.~Xu. 2019.
\newblock Deep unknown intent detection with margin loss.
\newblock \emph{ArXiv}, abs/1906.00434.

\bibitem[{Liu and Lane(2015)}]{liu2015recurrent}
Bing Liu and Ian Lane. 2015.
\newblock Recurrent neural network structured output prediction for spoken
  language understanding.
\newblock In \emph{Proc. NIPS Workshop on Machine Learning for Spoken Language
  Understanding and Interactions}.

\bibitem[{Liu and Lane(2016)}]{Liu_2016}
Bing Liu and Ian Lane. 2016.
\newblock Attention-based recurrent neural network models for joint intent
  detection and slot filling.
\newblock \emph{arXiv preprint arXiv:1609.01454}.

\bibitem[{Louvan and Magnini(2020)}]{Louvan2020RecentNM}
Samuel Louvan and B.~Magnini. 2020.
\newblock Recent neural methods on slot filling and intent classification for
  task-oriented dialogue systems: A survey.
\newblock In \emph{COLING}.

\bibitem[{Mesnil et~al.(2015)Mesnil, Dauphin, Yao, Bengio, Deng, Hakkani-Tur,
  He, Heck, Tur, Yu, and Zweig}]{Mesnil2015UsingRN}
Gr{\'e}goire Mesnil, Yann Dauphin, Kaisheng Yao, Yoshua Bengio, Li~Deng,
  Dilek~Z. Hakkani-Tur, Xiaodong He, Larry Heck, Gokhan Tur, Dong Yu, and
  Geoffrey Zweig. 2015.
\newblock Using recurrent neural networks for slot filling in spoken language
  understanding.
\newblock \emph{IEEE/ACM Transactions on Audio, Speech, and Language
  Processing}, 23:530--539.

\bibitem[{Ratinov and Roth(2009)}]{ratinov2009design}
Lev Ratinov and Dan Roth. 2009.
\newblock Design challenges and misconceptions in named entity recognition.
\newblock In \emph{Proceedings of the Thirteenth Conference on Computational
  Natural Language Learning (CoNLL-2009)}, pages 147--155.

\bibitem[{Ren et~al.(2019)Ren, Liu, Fertig, Snoek, Poplin, DePristo, Dillon,
  and Lakshminarayanan}]{Ren2019LikelihoodRF}
J.~Ren, Peter~J. Liu, E.~Fertig, Jasper Snoek, Ryan Poplin, Mark~A. DePristo,
  Joshua~V. Dillon, and Balaji Lakshminarayanan. 2019.
\newblock Likelihood ratios for out-of-distribution detection.
\newblock In \emph{NeurIPS}.

\bibitem[{Shah et~al.(2019)Shah, Gupta, Fayazi, and
  Hakkani-T{\"u}r}]{Shah2019RobustZC}
Darsh~J. Shah, Raghav Gupta, A.~Fayazi, and Dilek~Z. Hakkani-T{\"u}r. 2019.
\newblock Robust zero-shot cross-domain slot filling with example values.
\newblock \emph{ArXiv}, abs/1906.06870.

\bibitem[{Shu et~al.(2017)Shu, Xu, and Liu}]{Shu2017DOCDO}
Lei Shu, Hu~Xu, and Bing Liu. 2017.
\newblock Doc: Deep open classification of text documents.
\newblock \emph{ArXiv}, abs/1709.08716.

\bibitem[{Xu et~al.(2020{\natexlab{a}})Xu, He, Yan, hong Liu, Liu, and
  Xu}]{Xu2020ADG}
H.~Xu, Keqing He, Yuanmeng Yan, Si~hong Liu, Z.~Liu, and Weiran Xu.
  2020{\natexlab{a}}.
\newblock A deep generative distance-based classifier for out-of-domain
  detection with mahalanobis space.
\newblock In \emph{COLING}.

\bibitem[{Xu et~al.(2020{\natexlab{b}})Xu, He, Yan, Liu, Liu, and
  Xu}]{xu-etal-2020-deep}
Hong Xu, Keqing He, Yuanmeng Yan, Sihong Liu, Zijun Liu, and Weiran Xu.
  2020{\natexlab{b}}.
\newblock \href {https://www.aclweb.org/anthology/2020.coling-main.125} {A deep
  generative distance-based classifier for out-of-domain detection with
  mahalanobis space}.
\newblock In \emph{Proceedings of the 28th International Conference on
  Computational Linguistics}, pages 1452--1460, Barcelona, Spain (Online).
  International Committee on Computational Linguistics.

\bibitem[{Yan et~al.(2020)Yan, He, Xu, Liu, Meng, Hu, and
  Xu}]{yan-etal-2020-adversarial}
Yuanmeng Yan, Keqing He, Hong Xu, Sihong Liu, Fanyu Meng, Min Hu, and Weiran
  Xu. 2020.
\newblock \href {https://doi.org/10.18653/v1/2020.emnlp-main.490} {Adversarial
  semantic decoupling for recognizing open-vocabulary slots}.
\newblock In \emph{Proceedings of the 2020 Conference on Empirical Methods in
  Natural Language Processing (EMNLP)}, pages 6070--6075, Online. Association
  for Computational Linguistics.

\bibitem[{Yang and Mitchell(2017)}]{Yang2017LeveragingKB}
B.~Yang and Tom~Michael Mitchell. 2017.
\newblock Leveraging knowledge bases in lstms for improving machine reading.
\newblock In \emph{ACL}.

\bibitem[{Zeng et~al.(2021{\natexlab{a}})Zeng, He, Yan, Xu, and
  Xu}]{Zeng2021AdversarialSL}
Zhiyuan Zeng, Keqing He, Yuanmeng Yan, Hong Xu, and Weiran Xu.
  2021{\natexlab{a}}.
\newblock Adversarial self-supervised learning for out-of-domain detection.
\newblock In \emph{NAACL}.

\bibitem[{Zeng et~al.(2021{\natexlab{b}})Zeng, Xu, He, Yan, Liu, Liu, and
  Xu}]{9413908}
Zhiyuan Zeng, Hong Xu, Keqing He, Yuanmeng Yan, Sihong Liu, Zijun Liu, and
  Weiran Xu. 2021{\natexlab{b}}.
\newblock \href {https://doi.org/10.1109/ICASSP39728.2021.9413908} {Adversarial
  generative distance-based classifier for robust out-of-domain detection}.
\newblock In \emph{ICASSP 2021 - 2021 IEEE International Conference on
  Acoustics, Speech and Signal Processing (ICASSP)}, pages 7658--7662.

\bibitem[{Zhao and Feng(2018)}]{zhao-feng-2018-improving}
Lin Zhao and Zhe Feng. 2018.
\newblock \href {https://doi.org/10.18653/v1/P18-2068} {Improving slot filling
  in spoken language understanding with joint pointer and attention}.
\newblock In \emph{Proceedings of the 56th Annual Meeting of the Association
  for Computational Linguistics (Volume 2: Short Papers)}, pages 426--431,
  Melbourne, Australia. Association for Computational Linguistics.

\bibitem[{Zheng et~al.(2020)Zheng, Chen, and Huang}]{Zheng2020OutofDomainDF}
Yinhe Zheng, Guanyi Chen, and Minlie Huang. 2020.
\newblock Out-of-domain detection for natural language understanding in dialog
  systems.
\newblock \emph{IEEE/ACM Transactions on Audio, Speech, and Language
  Processing}, 28:1198--1209.

\end{thebibliography}


\end{document}


\maketitle
\appendix

\section{Implementation Details}
We use the public pre-trained Bert-large-uncased model to embed tokens which has 24 layers, 1024 hidden states, 16 headsand 336M parameters. The hidden size for the BiLSTM layer is set to 128. Adam is used for optimization with an initial learning rate of 1e-3. The dropout value is fixed at 0.5, and the batch size is 64. We trained the model only on in-domain labeled data. The training stage has an early stop setting with patience equal to 10. We use the best F1 scores on the validation set to calculate the MSP and GDA thresholds adaptively. Each result of the experiments is tested 10 times under the same setting and gets the average value. The training stage of our model lasts about 28 minutes on single Tesla T4 GPU(16 GB of memory).